\documentclass[10pt,twocolumn,letterpaper]{article}

\usepackage[pagenumbers]{cvpr}

\usepackage{xcolor, colortbl}
\usepackage{adjustbox} 
\usepackage{booktabs}
\usepackage{siunitx}

\usepackage[utf8]{inputenc} 
\usepackage[T1]{fontenc}    
\usepackage{url}            
\usepackage{booktabs}       
\usepackage{amsfonts}       
\usepackage{nicefrac}       
\usepackage{microtype}      
\usepackage{xcolor}         
\usepackage{microtype}
\usepackage{graphicx}
\usepackage{subcaption}
\usepackage{natbib}
\usepackage{cuted}   
\usepackage{capt-of} 

\usepackage{booktabs} 

\usepackage{amsmath}
\usepackage{amssymb}
\usepackage{mathtools}
\usepackage{amsthm}
\usepackage{bm}
\usepackage{booktabs}
\usepackage{multirow}

\usepackage{graphicx}      
\usepackage{caption}       
\usepackage{float}         
\usepackage{algorithm}     
\usepackage{algpseudocode} 

\theoremstyle{plain}

\theoremstyle{definition}

\theoremstyle{remark}

\usepackage[textsize=tiny]{todonotes}

\usepackage{xurl}
\usepackage{xcolor}
\usepackage{soul}
\usepackage[utf8]{inputenc}
\usepackage{subcaption}

\usepackage{multirow}
\usepackage{makecell}
\usepackage{array}
\usepackage{latexsym}
\usepackage{xspace}
\usepackage{mathrsfs}
\usepackage{accents}
\usepackage{caption}

\newcommand{\baby}{L\textsc{lp-dc}\xspace}
\newcommand{\eat}[1]{}

\definecolor{cvprblue}{rgb}{0.21,0.49,0.74}
\usepackage[pagebackref,breaklinks,colorlinks,allcolors=cvprblue]{hyperref}

\def\paperID{*****} 
\def\confName{CVPR}
\def\confYear{2026}

\title{Learning from Label Proportions with Dual-proportion Constraints}

\author{
Tianhao Ma\textsuperscript{1,2,3}, 
Ximing Li\textsuperscript{2,3,4}\thanks{Corresponding author.}, 
Changchun Li\textsuperscript{2,3}, 
Renchu Guan\textsuperscript{2,3}\vspace{2pt}\\
\textsuperscript{1}The University of Tokyo, Japan\\
\textsuperscript{2}College of Computer Science and Technology, Jilin University, China\\
\textsuperscript{3}Key Laboratory of Symbolic Computation and Knowledge Engineering of Ministry of Education,\\ Jilin University, China\\
\textsuperscript{4}RIKEN Center for Advanced Intelligence Project, Japan\\
{
    \tt\small \{matianhao2120, liximing86, changchunli93\}@gmail.com, guanrenchu@jlu.edu.cn
}
}
\begin{document}
\maketitle
\begin{abstract}
Learning from Label Proportions (LLP) is a weakly supervised problem in which the training data comprise bags, that is, groups of instances, each annotated only with bag-level class label proportions, and the objective is to learn a classifier that predicts instance-level labels. This setting is widely applicable when privacy constraints limit access to instance-level annotations or when fine-grained labeling is costly or impractical. In this work, we introduce a method that leverages Dual proportion Constraints (LLP-DC) during training, enforcing them at both the bag and instance levels. Specifically, the bag-level training aligns the mean prediction with the given proportion, and the instance-level training aligns hard pseudo-labels that satisfy the proportion constraint, where a minimum-cost maximum-flow algorithm is used to generate hard pseudo-labels. Extensive experimental results across various benchmark datasets empirically validate that LLP-DC consistently improves over previous LLP methods across datasets and bag sizes. The code is publicly available at \url{https://github.com/TianhaoMa5/CVPR2026_Findings_LLP_DC}.

\end{abstract}    
\section{Introduction}
Fueled by large-scale annotations and increasingly powerful architectures, fully supervised learning has driven rapid progress across core vision tasks~\citep{lin2025mvportrait,lin2025interanimate,lin2025creative4u}. In image classification, deep convolutional networks and transformers have established strong baselines and sustained state-of-the-art improvements~\citep{Krizhevsky2012ImageNet,Dosovitskiy2021ViT}. Object detection has likewise advanced through region-based frameworks and one-stage detectors, enabling accurate and efficient localization~\citep{Ren2015FasterRCNN,He2017MaskRCNN,Redmon2016YOLOv2}. Despite these successes, the substantial cost of dense, clean labels motivates studying alternatives to full supervision. 
In this context, \textbf{W}eakly-\textbf{S}upervised (\textbf{WS}) learning refers to paradigms with incomplete, inexact, and inaccurate supervision~\citep{zhou2018brief,book:Sugiyama+etal:2022}. During the past decades, many specific WS learning tasks have been widely investigated, including positive and unlabeled learning~\citep{guo2020positive,niu2016theoretical,Li2024}, partial label learning~\citep{wang2025rethinking,lv2020progressive,li2023learning}, and noisy label learning~\citep{li2021provably,Kim2024}, to name just a few.

\begin{figure}[t]
  \centering
  \includegraphics[width=\columnwidth]{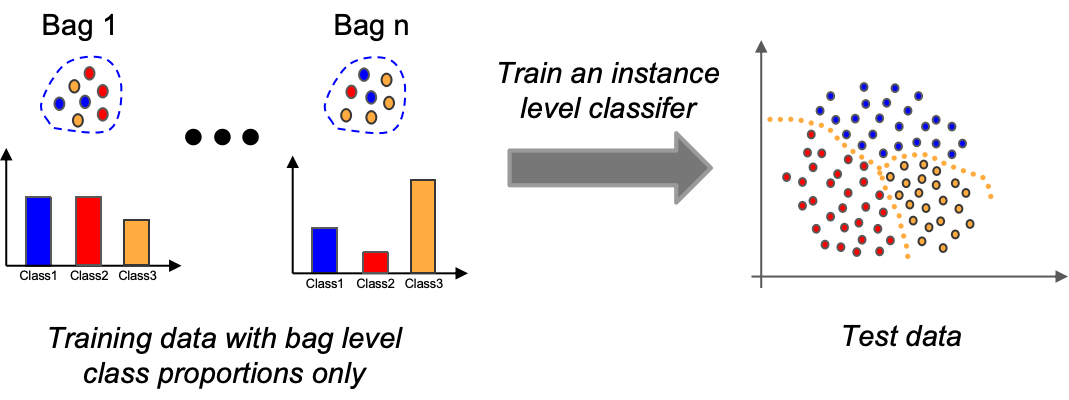}
  \caption{Training an instance-level classifier from a dataset with only bag-level class proportions.}
  \label{fig:llp-intro}
\end{figure}

A branch of WS learning deals with classification tasks, where groups of instances, called bags, with aggregate supervision are available only for privacy-preserving reasons~\citep{apple2023skadnetwork,diemert2022lessons}. \textbf{L}earning from \textbf{L}abel \textbf{P}roportions (\textbf{LLP}) is a special case in which bags are annotated with bag-level class proportions, while the goal is to train a classifier that predicts instance-level labels~\citep{asanomi2023mixbag,havaldar2023learning,brahmbhatt2023pac}, as illustrated in Fig.~\ref{fig:llp-intro}.
For example, real-world applications of LLP span diverse domains, including video event detection~\citep{lai2014video}, remote sensing~\citep{ding2017learning}, and disease diagnosis~\citep{tokunaga2020negative}.

During the past decades, many LLP methods have been proposed. A straightforward solution is to directly fit the mean bag-level predictions with the given class proportions~\citep{yu2014on,ardehaly2017co}. Recent mainstream LLP methods upgrade it by simultaneously fitting instance-level pseudo-labels in a self-training manner~\citep{dulac-arnold2019deep,liu2021two,liu2022self-llp,ma2024auxiliary,matsuo2023Learning,havaldar2023learning}. For example, pseudo-labels may be derived directly from model predictions~\citep{liu2022self-llp,ma2024auxiliary,matsuo2023Learning}, generated via Gibbs sampling~\citep{havaldar2023learning}, or constructed using optimal transport~\citep{liu2021two,dulac-arnold2019deep}. Additionally, some other studies focus on robust loss functions with solid theoretical guarantees~\citep{Li2024,busa2025nearly,zhang2022learning,busa-fekete2023easy}. However, these methods exhibit certain limitations in practical applications, such as producing negative loss values~\citep{busa-fekete2023easy} or being restricted to binary classification~\citep{busa2025nearly}.

In this paper, we introduce \textbf{\baby}, a novel method that, during training, simultaneously applies proportion constraints at the bag level and the instance level, with an emphasis on practical performance.
 At the bag-level, we follow prior work by using the given proportion as the target for the mean prediction of each bag; at the instance-level, we reinterpret LLP as a candidate label assignment problem. Based on this formulation, we use the current model outputs and solve a minimum-cost maximum-flow problem to efficiently assign hard pseudo-labels that satisfy the bag-level proportion constraints with the highest probability. Our method differs from approaches based on optimizing a relaxed optimal transport~\citep{dulac-arnold2019deep,liu2021two}, which iteratively generate soft labels under proportion constraints. Extensive experiments on various benchmark datasets demonstrate that \baby consistently outperforms existing LLP baselines.

In a nutshell, the contributions of this paper are listed as follows:
\begin{itemize}
    \item We propose \textbf{\baby}, a novel LLP method that efficiently generates pseudo-labels consistent with the label-proportion constraint for instance-level training.
    \item We conduct extensive experiments on various benchmark datasets, and empirical results indicate the effectiveness of \baby.
\end{itemize}
\section{Related Work}
During the past decades, many LLP methods have been widely investigated for binary and multi-class classification ~\citep{Quadrianto2008Estimating, krizhevsky2009learning, kueck2012learning, musicant2007supervised,yu2013proptosvm,ruping2010svm,patrini2014almost,ma2026learning}. Initially, a naive solution for LLP was to directly train our classifier to match the given bag–level label proportions, called empirical proportion risk minimization; and they mainly focus on binary classification ~\citep{yu2014on} while DLLP extends the method to multi-class classification ~\citep{ardehaly2017co}. However, these methods are inconsistent with the goal of predicting instance labels and heavily rely on the learning ability of the classifier, so they may empirically result in worse classification performance ~\citep{li2024optimistic,busa-fekete2023easy}. 
Recently, a line of work on LLP has introduced algorithms with strong theoretical guarantees~\citep{busa-fekete2023easy,li2024optimistic,busa2025nearly,zhang2022learning,saket2022algorithms,saket2021learnability}; however, these methods suffer from several practical limitations.
For example, ~\citep{busa-fekete2023easy} leveraged the fact that the expected proportion of each label in a bag is equal to the class prior distribution, however, this loss value can reach negative values, which are obviously improper, resulting in severe overfitting issues. 
~\citep{li2024optimistic} suggested a mean squared error-specific estimator, however, it is difficult to tune in practice and converges slowly during training.~\citep{wei2023universal,zhang2020learning} introduced a universal approach to aggregate-observation problems, including LLP; however, its computational complexity grows prohibitively with bag size. 

 Another branch of work focuses on improving practical performance~\citep{liu2021two,ma2024auxiliary,dulac-arnold2019deep,Liu2019Learning,havaldar2023learning,liu2022self-llp}. Some works incorporate representation learning modules such as generative adversarial networks ~\citep{Liu2019Learning}, consistency regularization~\citep{tsai2020learning}, and contrastive learning~\citep{yang2021two}. Mainstream methods regard the provided label proportions~\citep{yu2014on,ardehaly2017co} as supervision for bag-level mean predictions; in parallel, model predictions are leveraged to construct pseudo-labels for instance-level training. ~\citep{ma2024auxiliary} generates pseudo-labels directly from model predictions and weights them by prediction entropy, whereas ~\citep{liu2022self-llp} derives pseudo-labels by aggregating predictions over the entire training trajectory. In contrast, ~\citep{havaldar2023learning} employs a Gibbs model with belief propagation to obtain pseudo-labels; however, although these methods incorporate proportion information during pseudo-label generation, they do not strictly satisfy the bag-level proportion constraints. OPL~\citep{matsuo2023Learning} generates hard pseudo-labels satisfying bag proportions by solving a constrained optimization problem with cumulative unlikelihood.
~\citep{dulac-arnold2019deep,liu2021two} iteratively apply optimal transport to obtain soft pseudo-labels that approximately satisfy bag-level proportions; in contrast, \baby yields hard labels that strictly satisfy the proportion constraints, providing explicit instance-level supervision and promoting confident predictions with sharper decision boundaries. In the experimental section, we present a detailed comparison of these approaches, covering accuracy and wall-clock training time.

Additionally, there are some interesting studies that support various data generation processes for LLP data ~\citep{zhang2022learning, saket2021learnability, saket2022algorithms}. For example, ~\citep{saket2022algorithms,saket2021learnability} assume that an instance can belong to multiple bags. In our work, we focus on the most common practical case where instances are independent and randomly grouped into bags.

\section{Method}
In this section, we introduce the proposed LLP method named \baby.
\paragraph{Formulation of LLP}
Formally, let $\mathcal{X}\subset\mathbb{R}^d$ be the feature space and $\mathcal{Y}=\{1,\dots,l\}$ the label set, where $l$ is the number of classes. We define a classifier $f\in\mathcal{F}$ with $f:\mathcal{X}\to\mathbb{R}^l$; the goal is to learn an instance-level classifier when only bag-level proportions are observed. We assume that $\{(\mathbf{x}_i,y_i)\}_{i=1}^m$ are \textit{i.i.d.} samples from the data distribution, but the ground-truth labels $y_i$ are unobserved. Each labeled bag is denoted by $(\mathbf{B},\bm{\alpha})$, where $\mathbf{B}=(\mathbf{x}_1,\ldots,\mathbf{x}_m)$ contains the instance features and $\bm{\alpha}\in\Delta^{\,l-1}$ is the vector of class proportions defined as
\(
\alpha_c=\frac{1}{m}\sum_{i=1}^m \mathbb{I}(y_i=c),
\)
with $m$ the bag size. Following prior work~\citep{busa2025nearly,ma2024auxiliary}, we assume all bags have equal size for notational simplicity; however, our method readily extends to settings with variable bag sizes.
 The dataset is thus
\(
\mathcal{D}=\{(\mathbf{B}_i,\bm{\alpha}_i)\}_{i=1}^n.
\)

We define
$
g(x) = \mathrm{softmax}(f(x)),
$
which yields a probability vector $g(x) \in \Delta^{l-1}$ that can be interpreted as the posterior distribution over classes induced by the model $f$.

\subsection{Pseudo-Labels Generation under Label Proportion Constraints}

\label{core_algorithm}
\subsubsection{LLP from Label Assignment Perspective}
In our work, we consider label proportions from a different perspective: there exists a unique multiset\footnote{A multiset is a generalized set that allows multiple occurrences of the same element.} of instance labels consistent with the given proportion $\bm{\alpha}$, which we denote by $\mathcal{M}(\bm{\alpha}, m)$. Formally,
\[
\mathcal{M}(\bm{\alpha}, m) 
= \{\, \underbrace{1,\dots,1}_{m \bm{\alpha}_1 \text{ times}}, 
   \underbrace{2,\dots,2}_{m \bm{\alpha}_2 \text{ times}}, 
   \dots, 
   \underbrace{l,\dots,l}_{m \bm{\alpha}_l \text{ times}} \,\},
\]
where $m$ is the bag size and each label $k \in \mathcal{Y}$ appears exactly $m \bm{\alpha}_k$ times. Correspondingly, the set of candidate label assignments is defined as
\[
\mathcal{Y}(\bm{\alpha}, m) 
= \Bigl\{ \hat{\mathbf{y}}=(\hat{y}_1,\dots,\hat{y}_m) \in \mathcal{Y}^m \;\Big|\; 
\frac{1}{m}\sum_{j=1}^m \bm{e}^{\hat{y}_j}  = \bm{\alpha} \Bigr\},
\]
here  $\bm{e}^{\hat{y}_j}$ denotes the one-hot  vector for class $\hat y_j$.
Each $\hat{\mathbf{y}}\in\mathcal{Y}(\bm{\alpha}, m)$ represents one possible way 
to assign the labels from $\mathcal{M}(\bm{\alpha}, m)$ to the $m$ instances. Since the ground-truth labels of individual instances are unobservable, the true label assignment underlying each bag remains unknown.
Under the assumption that instances within a bag are generated independently,
the posterior probability of a label assignment
$\hat{\mathbf{y}}=(\hat{y}_1,\dots,\hat{y}_m)\in\mathcal{Y}(\bm{\alpha},m)$ is given by
$$
p(\hat{\mathbf{y}}\mid \mathbf{B}) 
= \prod_{j=1}^m p(\hat{y}_j \mid \mathbf{x}_j),
$$
where $p(\hat{y}_j \mid \mathbf{x}_j)$ represents the  posterior probability of label $\hat{y}_j$ for instance $\mathbf{x}_j$.
\subsubsection{Pseudo-Labels Generation}
\begin{figure}[t]
  \centering
  \includegraphics[width=\columnwidth]{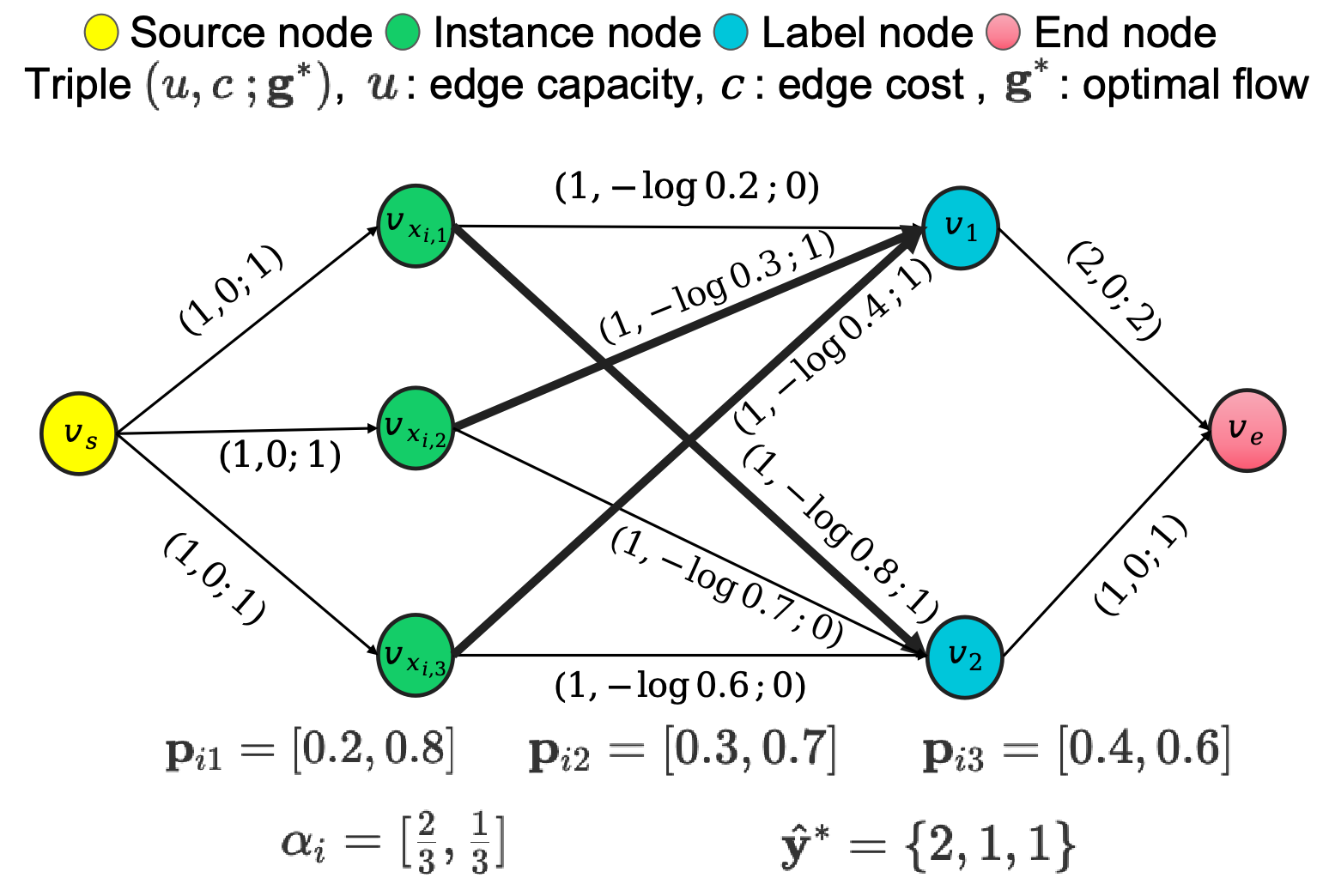}
  \caption{A toy example of minimum-cost maximum-flow problem with 3 instances and 2 labels. Given a labeled bag $\{\mathbf{x}_{i1},\mathbf{x}_{i2},\mathbf{x}_{i3},\bm{\alpha}_i\}$ and the current predicted outputs of instances $\{\mathbf{p}_{i1},\mathbf{p}_{i2},\mathbf{p}_{i3}\}$, we can form a corresponding directed multistage graph, and apply any off-the-shelf tool to solve for the optimal flow $\mathbf{g}^*$. It corresponds to the optimal candidate label assignment $\mathbf{\hat y}^*$ with the highest probability, indicated by the bold amounts for the bold edges between instance and label nodes. Best viewed in color.}
  \label{fig:mcmf}
\end{figure}
Our strategy is to use the highest probability label assignment derived from the model output and treat it as the pseudo-label for each instance. However, the size of $\mathcal{Y}(\bm{\alpha},m)$ grows combinatorially with the bag size $m$,
making the use of enumeration or sorting-based algorithms infeasible.

To solve this problem, we replace the enumeration problem spending at worst $O(m!)$ time cost with an efficient minimum-cost maximum-flow problem spending $O(m^2l^2)$ time cost.

Specifically, for each labeled bag $(\mathbf{B}_i,\bm{\alpha}_i)$, we form a corresponding directed multistage graph $\mathbf{G}=\{\mathcal{V},\mathcal{E},\mathbf{u},\mathbf{c}\}$, described as follows:

\paragraph{Vertex set $\mathcal{V}$} includes 4 types of nodes: a source node $v_s$, instance nodes $\{v_{\mathbf{x}_{ik}}\}_{k=1}^m$, label nodes $\{v_{y}\}_{y=1}^l$, and an end node $v_e$.

\paragraph{Edge set $\mathcal{E}$} includes 3 types of edges: edges from $v_s$ to $\{v_{\mathbf{x}_{ik}}\}_{k=1}^m$, edges from $\{v_{\mathbf{x}_{ik}}\}_{k=1}^m$ to $\{v_{y}\}_{y=1}^l$, and edges from $\{v_{y}\}_{y=1}^l$ to $v_e$.

\paragraph{Edge capacity $\mathbf{u}$} include the capacity constraint for each edge: $\{\mathbf{u}_{v_s \rightarrow v_{\mathbf{x}_{ik}}} = 1\}_{k=1}^m$, each $\{\mathbf{u}_{v_{\mathbf{x}_{ik}} \rightarrow v_{y}} = 1\}_{k=1,y=1}^{m,l}$, and each $\{\mathbf{u}_{v_{y} \rightarrow v_e} = \bm{\alpha}_{iy} \cdot m\}_{y=1}^l$.

\paragraph{Edge costs $\mathbf{c}$} include the unit cost for each edge: $\{\mathbf{c}_{v_s \rightarrow v_{\mathbf{x}_{ik}}} = 0\}_{k=1}^m$, $\{\mathbf{c}_{v_{\mathbf{x}_{ik}} \rightarrow v_{y}} = - \log p_{iky}\}_{k=1,y=1}^{m,l}$, and $\{\mathbf{c}_{v_{y} \rightarrow v_e} = 0\}_{y=1}^l$, where $p_{iky}$ denotes the current predicted output of $\mathbf{x}_{ik}$ belonging to label $y$, \ie higher predicted outputs correspond to lower unit costs of edges.

We define a flow $\mathbf{g}$ as an assignment scheme that assigns an amount to each edge satisfying the edge capacities $\mathbf{u}$. Our goal is to find an optimal flow $\mathbf{g}^*$ of minimum cost $\mathbf{c}^{\top}\mathbf{g}^*$ and maximum amount $\mathbf{u}^{\top}\mathbf{g}^*$ from the source node $v_s$ to the end node $v_e$. 

According to our graph setup, \textbf{the optimal flow $\mathbf{g}^*$ implies the optimal candidate label assignment $\hat{\mathbf{y}}_i^* 
= \arg\max_{\hat{\mathbf{y}}_i \in \mathcal{Y}^{m/ \bm{\alpha}_i}}
    \prod_{k=1}^m p_{ik\hat{y}_{ik}}
= \arg\min_{\hat{\mathbf{y}}_i \in \mathcal{Y}^{m/ \bm{\alpha}_i}}
    \sum_{k=1}^m -\log p_{ik\hat{y}_{ik}}$}. \textbf{First}, the amounts corresponding to the edges from $\{v_{\mathbf{x}_{ik}}\}_{k=1}^m$ to $\{v_{y}\}_{y=1}^l$ represent label assignments for instances. For each instance node $v_{\mathbf{x}_{ik}}$, it can be assigned by a single label because the edge capacities $\mathbf{u}_{v_s \rightarrow v_{\mathbf{x}_{ik}}}$ and $\{\mathbf{u}_{v_{\mathbf{x}_{ik}} \rightarrow v_{y}} = 1\}_{y=1}^{l}$ are equal to 1. \textbf{Second}, the minimum cost of $\mathbf{c}^{\top}\mathbf{g}^*$ corresponds to the label assignments with the highest probability because the edge costs $\{\mathbf{c}_{v_s \rightarrow v_{\mathbf{x}_{ik}}}\}_{k=1}^m$ and $\{\mathbf{c}_{v_{y} \rightarrow v_e}\}_{y=1}^l$ are equal to 0. \textbf{Third}, the amounts corresponding to the edges from $\{v_{y}\}_{y=1}^l$ to $v_e$ represent the number of assigned labels. The maximum amount $\mathbf{u}^{\top}\mathbf{g}^*$ indirectly satisfies the label proportion $\bm{\alpha}_i$ due to the edge capacities $\{\mathbf{u}_{v_{y} \rightarrow v_e} = \bm{\alpha}_{iy} \cdot m\}_{y=1}^l$. To solve for $\mathbf{g}^*$ (\ie $\mathbf{\hat y}_i^*$), we can apply any off-the-shelf minimum-cost maximum-flow algorithm\footnote{\url{https://developers.google.com/optimization}} with $O(m^2l^2)$ time cost ~\citep{ahuja1993network}. We present a toy example in Fig.~\ref{fig:mcmf} to illustrate the problem clearly, with the detailed computation provided in the experimental section.

\subsection{LLP with Dual-proportion Constraints}
\begin{figure*}[t]
    \centering
    \includegraphics[width=1\textwidth]{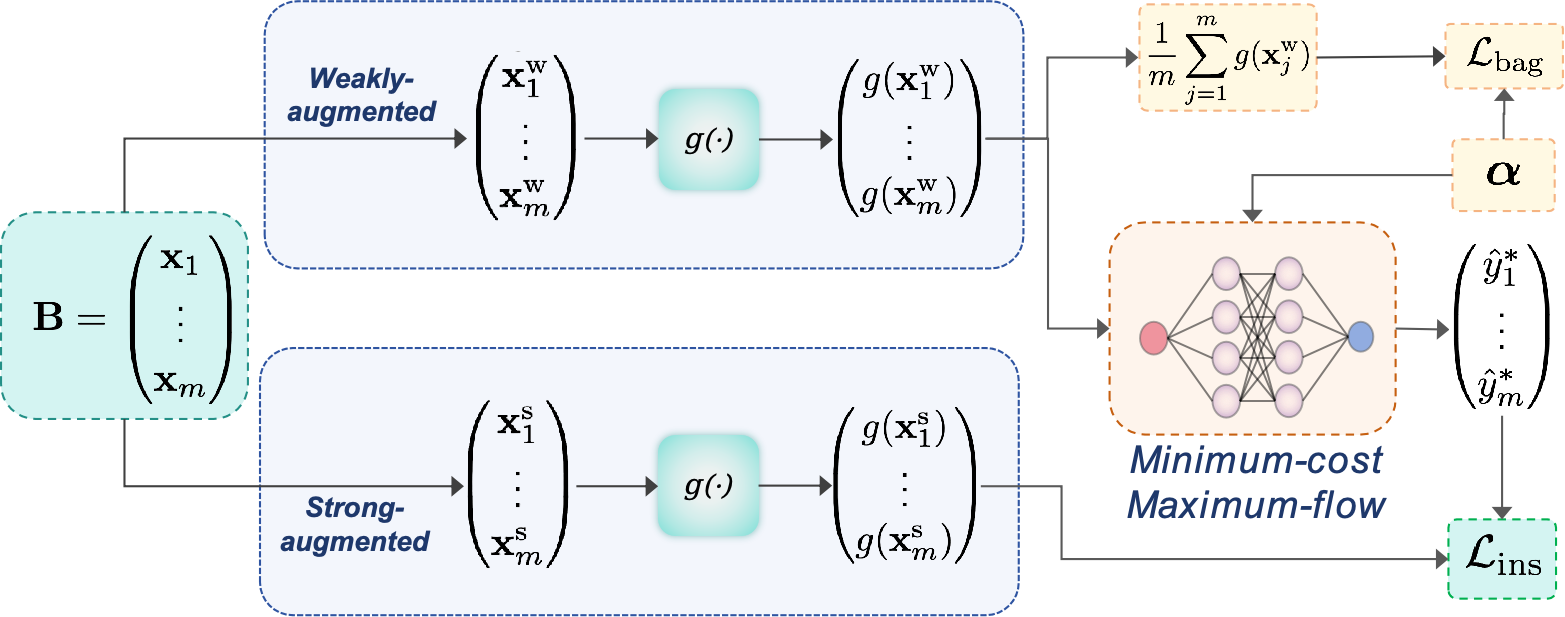} 
\caption{
Overview of \textbf{\baby}: the illustrated pipeline depicts the training process for a single bag.
Given weakly augmented instances $\big(\mathbf{x}_1^{\mathrm{w}},\dots,\mathbf{x}_m^{\mathrm{w}}\big)$ and strongly augmented instances $\big(\mathbf{x}_1^{\mathrm{s}},\dots,\mathbf{x}_m^{\mathrm{s}}\big)$, the model $g(\cdot)$ predicts on the weak views $g(\mathbf{x}_1^{\mathrm{w}}),\dots,g(\mathbf{x}_m^{\mathrm{w}})$; their average is aligned with the label proportion $\bm{\alpha}$ to compute the bag-level loss $\mathcal{L}_{\mathrm{bag}}$. 
Using these weak predictions and $\bm{\alpha}$, we build a graph and solve a minimum-cost maximum-flow problem to obtain pseudo-labels $\mathbf{y}^{*}$ that satisfy the proportion constraints. 
Finally, we compute the instance-level loss $\mathcal{L}_{\mathrm{ins}}$ on the strong views $g(\mathbf{x}_1^{\mathrm{s}}),\dots,g(\mathbf{x}_m^{\mathrm{s}})$ against $\mathbf{y}^{*}$.
}\label{Fig:pipeline}
\end{figure*}
Following~\citep{Sohn2020FixMatch,ma2024auxiliary}, we employ the widely used weak–strong augmentation strategy at the instance-level to enhance model robustness~\citep{li2024towards}. For each instance \(\mathbf{x}\), we denote by \(\mathbf{x}^{\mathrm{w}}\) the weakly augmented view and by \(\mathbf{x}^{\mathrm{s}}\) the strongly augmented view.

\paragraph{Bag-level Loss}
We introduce a bag-level proportion-consistency objective that aligns the average prediction of model with the target proportion~\citep{yu2014on,ardehaly2017co}:
\begin{equation}
\label{Eq2-2}
\mathcal{L}_{\mathrm{bag}} = \sum_{i=1}^n\ell\left(\frac{1}{m}\sum_{j=1}^m g(\mathbf{x}_{ij}^\mathrm{w}), \bm{\alpha}_i \right),
\end{equation}
where $\ell$ denotes a surrogate loss.
\paragraph{Instance-level Loss}
The pseudo-labels are generated from the model’s predictions on the weakly augmented inputs \(p_{ij}=g(\mathbf{x}_{ij}^{\mathrm{w}})\).
Using the algorithm introduced in the previous section, we derive \(\hat{{y}}^{*}_{ij}\) for each instance \(\mathbf{x}^\mathrm{w}_{ij}\) and employ it as the supervision target for the corresponding strongly augmented sample.
To mitigate the noise caused by unreliable pseudo-labels in the early stage of training, we set a fixed confidence threshold $\tau$. The instance-level loss is defined as follows:
\begin{equation}
\label{Eq2-3}
\mathcal{L}_{\mathrm{ins}}
= \sum_{i=1}^{n} \sum_{j=1}^{m}
\mathbb{I}(p_{ij\hat{y}^{*}_{ij}} \ge \tau)\,
\ell\!\left(g(\mathbf{x}_{ij}^{\mathrm{s}}),\, \hat{y}^{*}_{ij}\right),
\end{equation}
and the total loss is defined as follows:
\begin{equation}
\label{Eq2-4}
\mathcal{L} \;=\; \mathcal{L}_{\mathrm{bag}} \;+\; \lambda\,\mathcal{L}_{\mathrm{ins}},
\end{equation}
where $\lambda>0$ is a trade-off coefficient that balances the two terms. In implementation, we use the cross-entropy loss for $\ell$.

\paragraph{Training Summary of \baby}
We now show the full training process for the loss Eq.\eqref{Eq2-4}. Following previous studies~\citep{busa2025nearly,busa-fekete2023easy}, we apply the stochastic optimization method, which can efficiently handle large-scale LLP data. At each iteration $t$, we randomly load a mini-batch of labeled bags $\Omega^{(t)} = \{(\mathbf{B}_i,\bm{\alpha}_i)\}_{i=1}^{|\Omega^{(t)}|}$. For each labeled bag \((\mathbf{B}_i,\bm{\alpha}_i)\), we first apply the minimum-cost maximum-flow algorithm to obtain an optimal candidate label assignment \(\mathbf{g}_i^{*}\). We then compute the bag-level loss and the instance-level loss according to Eq.\eqref{Eq2-2} and Eq.\eqref{Eq2-3}, respectively. Finally, these losses are used to form stochastic gradients for Eq.\eqref{Eq2-4} to train the model. For clarity and completeness, \textbf{Algorithm 1} specifies the training process end to end, and Fig.~\ref{Fig:pipeline} offers a visual overview.

\section{Experiments}

\begin{algorithm}[t]
\caption{Training Procedure of \baby}
\label{alg:fairness}
\begin{algorithmic}[1]
\Require Training dataset $\{(\mathbf{B}_i, \bm{\alpha}_i)\}_{i=1}^n$; maximum number of epochs $T$.
\State Initialize parameters of the classifier $f$, set iteration number $t \gets 1$.
\While{not converged \textbf{and} $t \le T$} 
    \State Sample a mini-batch of labeled bags $\Omega^{(t)} = \{(\mathbf{B}_i, \bm{\alpha}_i)\}$.
    \State Generate two different augmented views for each instance.
    \For{each labeled bag $(\mathbf{B}_i, \bm{\alpha}_i) \in \Omega^{(t)}$}
    \State Compute the optimal candidate label assignment $\hat{\mathbf{y}}_i^*$ via the minimum-cost maximum-flow algorithm.
\EndFor
\State Compute the bag-level loss according to Eq.~\eqref{Eq2-2} and the instance-level loss according to Eq.~\eqref{Eq2-3}.
\State Form stochastic gradients based on Eq.~\eqref{Eq2-4}.

    \State Update the classifier $f$ using the SGD optimizer.
    \State $t \gets t + 1$
\EndWhile
\end{algorithmic}
\end{algorithm}

\begin{table*}[ht]
    \caption{Classification accuracy (mean ± std) on CIFAR-10, CIFAR-100, SVHN, Fashion-MNIST and MiniImageNet for different bag sizes. The highest accuracy is highlighted in bold. The symbol ``*'' indicates results for FLMm are reproduced directly from its original publication~\citep{yang2021two}, while the dash symbol ``-'' signifies the absence of reported results for SVHN, Fashion-MNIST, and MiniImageNet in FLMm's experiments.}
    \label{tab:result}
    \centering
    \fontsize{9.4pt}{9.4pt}\selectfont
    {%
    \renewcommand{\arraystretch}{1.25}
    \begin{tabular}{clccccc}
    \toprule
    \multicolumn{1}{c}{\multirow{2}[1]{*}{Dataset}} & \multicolumn{1}{c}{\multirow{2}[1]{*}{Model}} & \multicolumn{4}{c}{Bag Size} & \multirow{2}[1]{*}{{\parbox{1.5cm}{\centering Fully\\Supervised}}}  \\
\cmidrule{3-6}      &  & \multicolumn{1}{c}{16} & \multicolumn{1}{c}{32} & \multicolumn{1}{c}{64} & \multicolumn{1}{c}{128} \\
    \midrule
    \multicolumn{1}{c}{\multirow{7}[1]{*}{CIFAR-10}} 
      & DLLP & 91.59 ± 1.52 & 88.61 ± 0.90 & 79.76 ± 1.45 & 64.95 ± 0.01 & \multicolumn{1}{c}{\multirow{7}[1]{*}{96.05 ± 0.33}} \\
      & LLP-VAT & 91.80 ± 0.08 & 89.11 ± 0.22 & 78.75 ± 0.46 & 63.89 ± 0.19 \\
      & ROT & 94.86 ± 0.68 & 94.34 ± 0.65 & 93.97 ± 0.96 & 92.23 ± 0.81 \\
      & SoftMatch & 95.24 ± 0.12 & 95.25 ± 0.14 & 94.23 ± 0.18 & 93.87 ± 0.22 \\
      & FLMm* & 92.34 & 92.00 & 91.74 & 91.54 \\
      & L$^2$\textsc{p-ahil} & 94.96 ± 0.13 & 95.00 ± 0.11 & 94.58 ± 0.21 & 93.64 ± 0.20 \\
      & \baby & \textbf{95.97 ± 0.03} & \textbf{95.90 ± 0.07} & \textbf{95.46 ± 0.03} & \textbf{94.47 ± 0.05} \\
    \midrule
    \multicolumn{1}{c}{\multirow{7}[1]{*}{CIFAR-100}} 
      & DLLP & 71.28 ± 1.56 & 69.92 ± 2.86 & 53.58 ± 1.60 & 25.86 ± 2.15 & \multicolumn{1}{c}{\multirow{7}[1]{*}{79.89 ± 0.14}} \\
      & LLP-VAT & 73.85 ± 0.22 & 71.62 ± 0.07 & 65.31 ± 0.33 & 37.36 ± 0.63 \\
      & ROT & 72.74 ± 0.08 & 69.31 ± 0.22 & 17.48 ± 0.86 & 11.02 ± 0.79 \\
      & SoftMatch & 80.14 ± 0.12 & 2.40 ± 0.15 & 2.04 ± 0.10 & 2.12 ± 0.13 \\
      & FLMm* & 66.16 & 65.59 & 64.07 & 61.25 \\
      & L$^2$\textsc{p-ahil} & 78.65 ± 0.28 & 77.30 ± 0.50 & 76.52 ± 0.23 & 72.21 ± 0.37 \\
      & \baby & \textbf{80.32 ± 0.10} & \textbf{79.85 ± 0.03} & \textbf{79.05 ± 0.19} & \textbf{73.29 ± 0.26} \\
    \midrule
    \multicolumn{1}{c}{\multirow{7}[1]{*}{SVHN}} 
      & DLLP & 96.90 ± 0.50 & 96.93 ± 0.23 & 96.64 ± 0.32 & 95.51 ± 0.04 & \multicolumn{1}{c}{\multirow{7}[1]{*}{97.77 ± 0.03}}\\
      & LLP-VAT & 96.88 ± 0.03 & 96.68 ± 0.01 & 96.38 ± 0.10 & 95.29 ± 0.17 \\
      & ROT & 95.54 ± 0.10 & 94.78 ± 0.13 & 96.75 ± 0.11 & 26.00 ± 0.43 \\
      & SoftMatch & 22.39 ± 0.11 & 19.68 ± 0.13 & 19.60 ± 0.12 & 19.64 ± 0.14 \\
      & FLMm* & \multicolumn{1}{c}{-} & \multicolumn{1}{c}{-} & \multicolumn{1}{c}{-} & \multicolumn{1}{c}{-} \\
      & L$^2$\textsc{p-ahil} & 97.91 ± 0.02 & 97.88 ± 0.01 & 97.74 ± 0.06 & 97.67 ± 0.17 \\
      & \baby & \textbf{98.01 ± 0.02} & \textbf{97.99 ± 0.04} & \textbf{97.97 ± 0.02} & \textbf{97.97 ± 0.07} \\
    \midrule
    \multicolumn{1}{c}{\multirow{7}[1]{*}{\parbox{1.25cm}{Fashion-\\ MNIST}}} 
      & DLLP & 94.20 ± 0.02 & 93.70 ± 0.39 & 93.18 ± 0.22 & 91.70 ± 0.21 & \multicolumn{1}{c}{\multirow{7}[1]{*}{96.39 ± 0.02}}\\
      & LLP-VAT & 94.69 ± 0.20 & 94.17 ± 0.16 & 93.25 ± 0.18 & 92.30 ± 0.13 \\
      & ROT & 94.25 ± 0.17 & 93.68 ± 0.22 & 92.53 ± 0.46 & 91.84 ± 0.19 \\
      & SoftMatch & 95.85 ± 0.22 & {95.86 ± 0.25} & 95.18 ± 0.21 & 94.73 ± 0.20 \\
      & FLMm* & \multicolumn{1}{c}{-} & \multicolumn{1}{c}{-} & \multicolumn{1}{c}{-} & \multicolumn{1}{c}{-} \\
      & L$^2$\textsc{p-ahil} & \textbf{96.93 ± 0.23} & 95.78 ± 0.15 & \textbf{95.27 ± 0.13} & 94.19 ± 0.14 \\
      & \baby & 95.90 ± 0.02 & \textbf{95.86 ± 0.06} & 95.19 ± 0.20 & \textbf{94.74 ± 0.07} \\
    \midrule
    \multicolumn{1}{c}{\multirow{7}[1]{*}{MiniImageNet}} 
      & DLLP & 64.53 ± 0.41 & 55.37 ± 0.38 & 27.57 ± 0.20 & 9.06 ± 0.14 & \multicolumn{1}{c}{\multirow{7}[1]{*}{73.95 ± 0.22}} \\
      & LLP-VAT & 64.17 ± 0.34 & 54.36 ± 0.29 & 30.96 ± 0.24 & 9.69 ± 0.17 \\
      & ROT & 67.02 ± 0.34 & 27.49 ± 0.38 & 6.01 ± 0.30 & 3.50 ± 0.10 \\
      & SoftMatch & 2.02 ± 0.23 & 1.86 ± 0.24 & 1.95 ± 0.20 & 1.72 ± 0.33 \\
      & FLMm* & \multicolumn{1}{c}{-} & \multicolumn{1}{c}{-} & \multicolumn{1}{c}{-} & \multicolumn{1}{c}{-} \\
      & L$^2$\textsc{p-ahil} & \textbf{70.26 ± 0.26} & \textbf{59.81 ± 0.21} & 37.51 ± 0.16 & 16.91 ± 0.15 \\
      & \baby & 66.90 ± 0.40 & 59.46 ± 0.11 & \textbf{38.64 ± 0.71} & \textbf{19.01 ± 0.40} \\
    \bottomrule
    \end{tabular}%
    }
    \vspace{-3pt}
\end{table*}

\subsection{Settings}
\paragraph{Datasets}
In the experiments, we employ five widely used benchmark datasets:
Fashion-MNIST (F-MNIST)\footnote{\url{https://github.com/zalandoresearch/fashion-mnist}} (28$\times$28 grayscale images of 10 clothing categories with 60k/10k train/test split),
CIFAR-10\footnote{\url{https://www.cs.toronto.edu/~kriz/cifar.html}} (32$\times$32 color natural images over 10 object classes),
CIFAR-100\footnote{\url{https://www.cs.toronto.edu/~kriz/cifar.html}} (32$\times$32 color natural images over 100 classes grouped into 20 superclasses),
SVHN\footnote{\url{https://ufldl.stanford.edu/housenumbers/}} (street-view house number digits in the wild with hundreds of thousands of labeled 32$\times$32 crops),
and mini-ImageNet\footnote{\url{https://opendatalab.org.cn/OpenDataLab/Mini-ImageNet}} (an 84$\times$84 subset of ImageNet with 100 classes, commonly used for few-shot learning).

For each dataset, we construct controlled LLP variants by first randomly shuffling the instances and then uniformly partitioning them into non-overlapping bags. The label proportion of each bag is computed from the original instance labels. We vary the bag size \(m \in \{16, 32, 64, 128\}\), yielding 20 controlled LLP datasets in total. This randomized shuffle-and-assemble scheme mirrors common industrial practice, and due to such generation, it leads to bags with very similar label proportions~\citep{Scott2020LLP}, while simultaneously making the learning problem considerably more challenging.   
\paragraph{Baselines}

We compare \baby with six existing LLP methods that have demonstrated strong empirical performance:
A): DLLP~\citep{ardehaly2017co,yu2014on}, a vanilla LLP approach that treats label proportions as targets for mean predictions;
B): LLP-VAT~\citep{tsai2020learning}, which builds on DLLP by adding a consistency-regularization term;
C): ROT~\citep{dulac-arnold2019deep}, which uses optimal transport to construct pseudo-labels; we set the number of Sinkhorn iterations to 3.
D): SoftMatch~\citep{Chen2023SoftMatch}, a classic semi-supervised method that we use to generate pseudo-labels and combine with DLLP as a baseline;
E): FLMm~\citep{yang2021two}, which employs a deeper network and contrastive learning for fine-tuning via FLMe;
F): L$^2$\textsc{p-ahil}~\citep{ma2024auxiliary}, which adopts a Dual Entropy Weighting (DEW) strategy to adjust the pseudo-label weights.
*): Fully-Supervised indicates the standard setting in which all instance-level labels are available for supervised training.
All baseline results are taken directly from~\citep{ma2024auxiliary}.
\paragraph{Implementation Details}
\label{Sec:details}

For SVHN, Fashion-MNIST, and CIFAR-10, we adopt the WRN-28-2~\citep{zagoruyko2016wide} architecture as the encoder, while WRN-28-8 is used for CIFAR-100. For MiniImageNet, we employ the ResNet-18 architecture. The classifier is implemented as a single linear layer.  
Each training step uses a batch size equal to the bag size multiplied by the number of bags, resulting in a total of 1024 samples per step.  
Model optimization is performed using Stochastic Gradient Descent (SGD)~\citep{polyak1964some} with a momentum of 0.9. The weight decay is set to 5e-4 for WRN-28-2, 1e-3 for WRN-28-8, and 1e-4 for ResNet-18.  
The initial learning rate is set to \( \eta_0 = 0.03 \) for all datasets, except for MiniImageNet, where it is set to \( \eta_0 = 0.05 \). A cosine learning rate decay schedule~\citep{loshchilov2017sgdr} is applied as \( \eta = \eta_0 \cos\left(\frac{7\pi k}{16K}\right) \), where \( k \) denotes the current training step and \( K \) is the total number of steps.  
We train all models for \( 1024 \) epochs. Weak data augmentations include random horizontal flipping and random cropping. For SVHN, following~\citep{Sohn2020FixMatch}, horizontal flips are replaced with random translations up to 12.5\% in both directions. Strong augmentations are performed using RandAugment~\citep{cubuk2020randaugment}.  
The configuration for the fully supervised setting remains consistent with the above setup.

We set the hyperparameters in \baby to \(\lambda = 0.5\) and \(\tau = 0.6\). These values are not claimed to be optimal, but they are sufficient to showcase the effectiveness of our approach (see Sec.~\ref{sec:sens}).
\subsection{Results and Analysis}
\begin{figure*}
    \centering
    \includegraphics[width=1\linewidth]{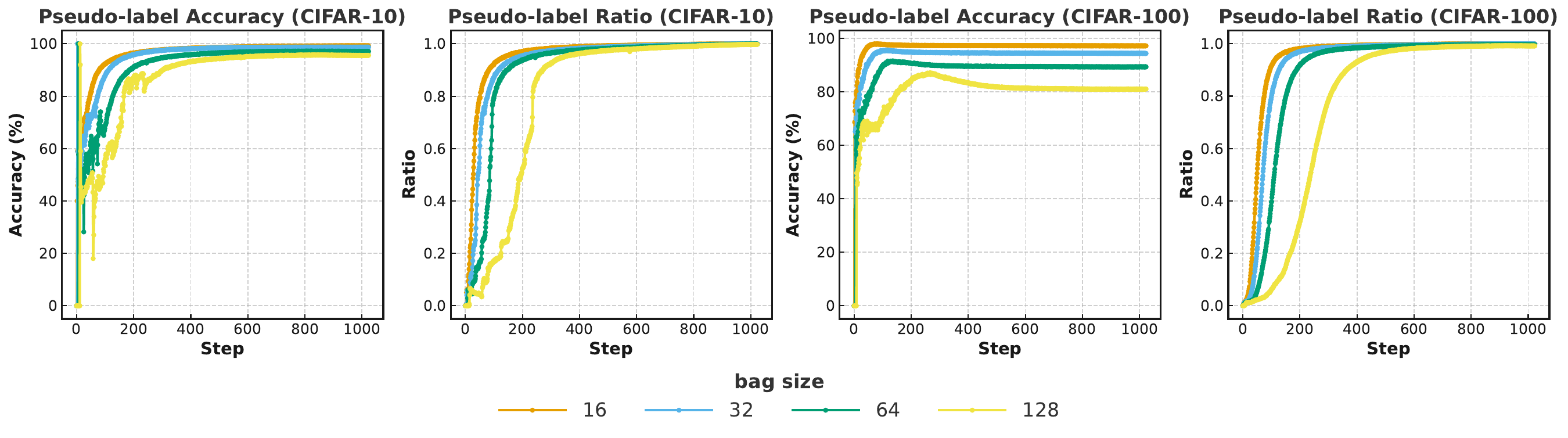}
    \caption{Pseudo-label accuracy (left) and pseudo-label ratio (right) during training on CIFAR-10 (left two panels) and CIFAR-100 (right two panels) under different bag sizes (16, 32, 64, 128).}
    \label{fig:placeholder}
\end{figure*}
\begin{figure*}
    \centering
    \includegraphics[width=1\linewidth]{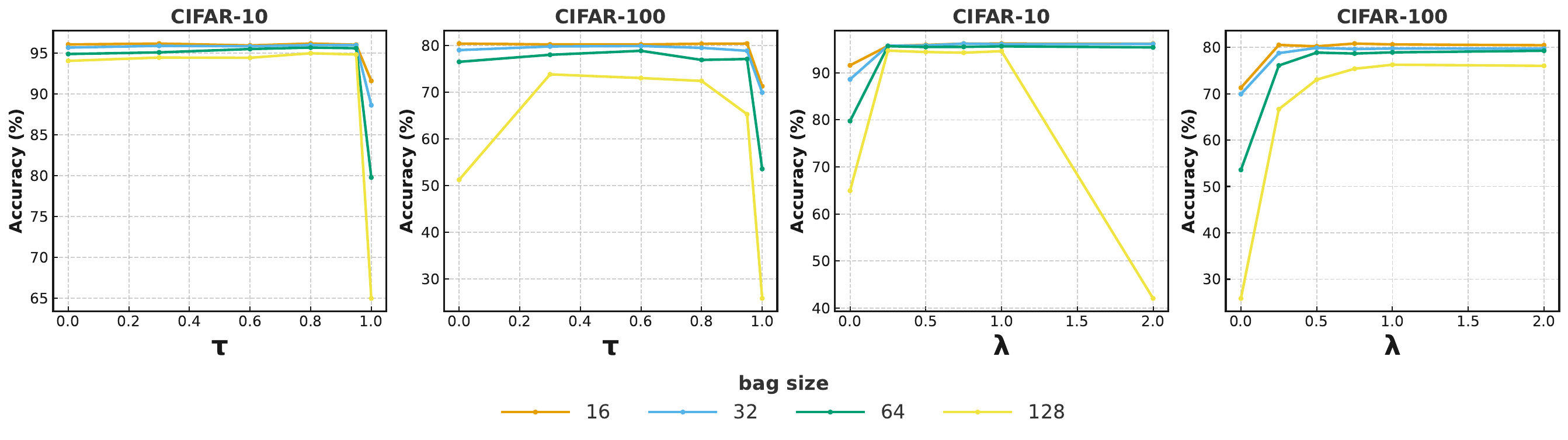}
    \caption{Sensitivity analysis of the instance-level loss weight \(\lambda\) in \(\mathcal{L}_{\mathrm{ins}}\) and the threshold $\tau$ parameter on CIFAR-10 and CIFAR-100. For \(\lambda \in \{2, 1, 0.75, 0.5, 0.25, 0\}\), performance remains stable over a moderate range, indicating robustness to the choice of \(\lambda\). Similarly, for $\tau$ \(\{0.95, 0.8, 0.6, 0.3, 0\}\), performance is consistent across a broad range, demonstrating robustness to threshold selection. Setting \(\lambda=0\) or the $\tau$ to 1 reduces the method to DLLP.
}\label{fig:tau_lambda}
\end{figure*}
As shown in Table~\ref{tab:result}, \baby consistently improves over previous LLP methods across datasets and bag sizes. The improvements are particularly noticeable on more challenging benchmarks such as CIFAR-100 and MiniImageNet.

\sisetup{
  table-number-alignment = center,
  table-figures-integer = 3,
  table-figures-decimal = 2,
  detect-weight=true,
}

\begin{table*}[!t]
  \centering
 \small
 \caption{Execution time results on {CIFAR100} and {MiniImageNet} (bag size $\in\{16,32,64,128\}$).
  Units are seconds per epoch. Reported numbers are the mean over 10 epochs under the same settings on a single NVIDIA A100 GPU.
  For ROT, {iter} denotes the number of Sinkhorn iterations.}
  \label{tab:runtime}
  \begin{tabular}{l *{4}{S} @{\hspace{18pt}} *{4}{S}}
    \toprule
    & \multicolumn{4}{c}{{CIFAR100}} & \multicolumn{4}{c}{{MiniImageNet}} \\
    \cmidrule{2-5}\cmidrule{6-9}
    \textbf{Method} & {16} & {32} & {64} & {128} & {16} & {32} & {64} & {128} \\
    \midrule
    DLLP            & 18.45 & 18.35 & 18.23 & 17.95 & 21.74 & 21.66 & 21.01 & 20.22 \\
    LLP-VAT         & 62.95 & 61.58 & 60.44 & 60.86 & 70.04 & 69.02 & 67.52 & 67.46 \\
    ROT (iter = $3$)& 40.63 & 38.50 & 36.97 & 36.65 & 47.23 & 43.91 & 43.09 & 42.39 \\
    ROT (iter = $75$)& 60.03 & 48.50 & 40.08 & 37.26 & 66.07 & 53.28 & 46.07 & 43.21 \\
    SoftMatch       & 39.45 & 38.70 & 36.45 & 36.40 & 46.03 & 42.95 & 42.23 & 41.73 \\
    L$^2$\textsc{p-ahil} & 39.67 & 39.53 & 36.86 & 36.54 & 46.86 & 43.46 & 42.91 & 41.58 \\
        \midrule

    \baby           & 56.60 & 53.93 & 52.65 & 51.43 & 63.01 & 62.65 & 59.32 & 58.32 \\
    \bottomrule
  \end{tabular}
\end{table*}
On CIFAR-10, \baby achieves the best performance across all bag sizes, reaching 95.97\% at bag size 16 and 94.47\% at bag size 128, outperforming both L$^2$\textsc{p-ahil} and SoftMatch. The performance remains stable even when the bag size increases and supervision becomes weaker.

For CIFAR-100, the advantage of \baby becomes more apparent. At bag size 16, it reaches 80.32\%, compared with 78.65\% for L$^2$\textsc{p-ahil} and 80.14\% for SoftMatch. As the bag size increases, several baselines degrade noticeably, while \baby maintains relatively strong performance (73.29\% at bag size 128).

On SVHN, \baby achieves 98.01\% at bag size 16, slightly improving over the previous best result of 97.91\% from L$^2$\textsc{p-ahil}. The method also maintains competitive results across all bag sizes.

For Fashion-MNIST, \baby obtains 95.90\% at bag size 16 and 94.74\% at bag size 128, performing comparably to or better than the strongest baselines under the same settings.

On MiniImageNet, which is more challenging and has fewer training samples, \baby shows clearer improvements for larger bags. In particular, it reaches 38.64\% and 19.01\% at bag sizes 64 and 128, improving over L$^2$\textsc{p-ahil} by 1.13\% and 2.10\%, respectively.

Overall, the results show that \baby performs consistently well across datasets and bag sizes.

\subsection{Training Curves for Pseudo-Labels}
We trained on CIFAR-10 and CIFAR-100 and tracked both pseudo-label accuracy and ratio over training steps for different bag sizes, as shown in Fig.~\ref{fig:placeholder}. The curves show a clear upward trend: accuracy climbs quickly and then levels off, while the ratio steadily approaches 1, meaning more samples receive confident pseudo-labels as training progresses. Smaller bags (16/32) ramp up faster and reach slightly higher final accuracy; larger bags (especially 128) rise more slowly and end a bit lower on CIFAR-100. CIFAR-10 shows minor early jitter but stabilizes quickly. Overall, pseudo-label quality and coverage improve steadily throughout training, with smaller bags giving faster and stronger convergence.

\subsection{Sensitivity Analysis}

\label{sec:sens}

\paragraph{Parameter $\tau$}
The threshold $\tau$ controls which pseudo-labels are used for instance-level supervision by filtering out low-confidence predictions. We evaluate different values of $\tau$ in the left panels of Fig.~\ref{fig:tau_lambda}. On CIFAR-100 with bag size 128, using a moderate threshold improves performance by removing unreliable pseudo-labels while still retaining sufficient training signals. When $\tau$ is set too high, very few pseudo-labels are selected and the model receives little instance-level supervision. Conversely, very small thresholds allow noisy labels to be used, which can harm performance. In practice, the method is not very sensitive to this parameter: the accuracy remains stable over a relatively wide range of values (approximately $0.3$–$0.9$). This behavior differs from semi-supervised approaches such as FixMatch~\citep{Sohn2020FixMatch}, which often rely on very high confidence thresholds.

\paragraph{Parameter $\lambda$}
The parameter $\lambda$ controls the contribution of the instance-level loss relative to the bag-level objective. We vary $\lambda$ in the right panels of Fig.~\ref{fig:tau_lambda}. When $\lambda$ is very small, the instance-level objective has little effect and the model relies mainly on bag-level supervision, which can weaken fine-grained discrimination. Increasing $\lambda$ improves performance by encouraging the model to utilize pseudo-labels for instance-level learning. However, overly large values may amplify errors in pseudo-label assignments and lead to unstable training. Empirically, the method performs well over a moderate range of $\lambda$, roughly between $0.5$ and $1$, without requiring dataset-specific tuning.

\subsection{Runtime Analysis}

We report the runtime comparison of different LLP methods in Table~\ref{tab:runtime}. 
Compared with the baseline DLLP, which uses only a single type of data augmentation and thus runs relatively fast, 
all other methods employ strong--weak augmentation and additional regularization to improve performance, increasing computation.
Nevertheless, our method \baby, despite introducing an extra pseudo-label generation step, 
exhibits comparable runtime to these approaches. 
For example, on CIFAR100 with bag sizes 16/128, \baby\ is faster than LLP-VAT by  6.35 s and 9.43 s per epoch, respectively; 
on MiniImageNet with bag sizes 32/64, the gaps are just 6.37 s and 8.20 s per epoch. 
These results indicate that the overhead from pseudo-label generation in \baby is negligible in practice. 
For ROT, increasing the number of Sinkhorn iterations makes the soft pseudo-labels better match the prescribed label proportions $\bm{\alpha}$, 
but also raises the computational cost (e.g., moving from 3 to 75 iterations increases epoch time across bag sizes), 
illustrating that tighter proportion adherence entails higher time complexity. 
In contrast, \baby\ uses a fixed and efficient pseudo-label generation/update procedure, achieving a favorable balance between accuracy and efficiency.

\section{Conclusion}
In this work we proposed \baby, whose core idea is to enforce label–proportion consistency at both the bag level and the instance level. At the bag level, the average of model predictions matches the given proportions. At the instance level, training uses pseudo-labels that strictly satisfy the same proportion constraints. We cast pseudo-label assignment as a minimum-cost maximum-flow problem, which enables fast generation of proportion-consistent labels at scale. Extensive experiments on standard benchmarks demonstrate the effectiveness of our approach, and comprehensive ablations and sensitivity studies indicate strong robustness with low hyperparameter sensitivity. 
Looking ahead, our proportion-aware pseudo-labeling can be combined with sample selection strategies such as adaptive thresholds, stage-aware curricula, and teacher–student frameworks to further refine pseudo-label quality while preserving proportion constraints.
In addition, because LLP currently lacks standardized real-world datasets for computer vision, partly due to privacy and related constraints, we plan to construct and release a benchmark suite to facilitate fair comparison and foster future research in this area.
{
    \small
    \bibliographystyle{ieeenat_fullname}
    \bibliography{LLP_MCMF}             
}


\end{document}